\PassOptionsToPackage{table}{xcolor} 
\documentclass[11pt]{article}

\usepackage[final]{acl}

\usepackage{times}
\usepackage{latexsym}

\usepackage[T1]{fontenc}

\usepackage[utf8]{inputenc}

\usepackage{microtype}

\usepackage{inconsolata}

\usepackage{multicol} 
\usepackage{graphicx}
\usepackage{booktabs} 
\usepackage{multirow} 
\usepackage{adjustbox} 
\usepackage{amssymb} 
\usepackage{tabularx} 
\usepackage[table]{xcolor} 
\usepackage{hyperref}  
\usepackage{marvosym}

\definecolor{lightgreen}{HTML}{E8F5E9}
\definecolor{lightorange}{HTML}{FFF3E0}
\definecolor{lightred}{HTML}{FFEBEE}
\definecolor{darkred}{HTML}{B71C1C}

\pdfoutput=1

\title{Visual Merit or Linguistic Crutch? A Close Look at DeepSeek-OCR}

\author{
  \textbf{Yunhao Liang$^{1,2}$\thanks{~Equal Contribution.} \quad}
  \textbf{Ruixuan Ying$^{3}$\footnotemark[1] \quad} 
  \textbf{Bo Li$^{4}$\quad}
  \textbf{Hong Li$^{4}$\quad}
  \textbf{Kai Yan$^{4}$\quad}
  \textbf{Feiteng Fang$^{7}$\quad}\\
  \textbf{Qingwen Li$^{4}$\quad}
  \textbf{Min Yang$^{7,8}$\quad}
  \textbf{Okamoto Satoshi$^{3,5,6}$\quad}
  \textbf{Zhe Cui$^{1,2}$\quad}
  \textbf{Shiwen Ni$^{7,8}$\thanks{~Corresponding author. \scalebox{0.8}{\Letter} \href{mailto:sw.ni@siat.ac.cn}{sw.ni@siat.ac.cn}}}
  \\ \\
  $^1$Chengdu Institute of Computer Applications, CAS ~ $^2$University of Chinese Academy of Sciences \\
  $^3$IMRAM, Tohoku University ~$^4$China Tower Corporation Limited ~$^5$CSIS, Tohoku University \\$^6$National Institute for Materials Science ~$^7$Shenzhen Institutes of Advanced Technology, CAS\\
  $^8$Artificial Intelligence Research Institute, Shenzhen University of Advanced Technology
}
\begin{document}
\maketitle
\begin{abstract}
DeepSeek-OCR utilizes an optical 2D mapping approach to achieve high-ratio vision-text compression, claiming to decode text tokens exceeding ten times the input visual tokens. While this suggests a promising solution for the LLM long-context bottleneck, we investigate a critical question: "Visual merit or linguistic crutch—which drives DeepSeek-OCR’s performance?" By employing sentence-level and word-level semantic corruption, we isolate the model's intrinsic OCR capabilities from its language priors. Results demonstrate that without linguistic support, DeepSeek-OCR's performance plummets from approximately 90\% to 20\%. Comparative benchmarking against 13 baseline models reveals that traditional pipeline OCR methods exhibit significantly higher robustness to such semantic perturbations than end-to-end methods. Furthermore, we find that lower visual token counts correlate with increased reliance on priors, exacerbating hallucination risks. Context stress testing also reveals a total model collapse around 10,000 text tokens, suggesting that current optical compression techniques may paradoxically aggravate the long-context bottleneck. This study empirically defines DeepSeek-OCR’s capability boundaries and offers essential insights for future optimizations of the vision-text compression paradigm. We release all data, results and scripts used in this study at \href{https://github.com/dududuck00/DeepSeekOCR}{github}.
\end{abstract}

\section{Introduction}
\begin{figure}[t]
    \centering
    \includegraphics[width=0.49\textwidth]{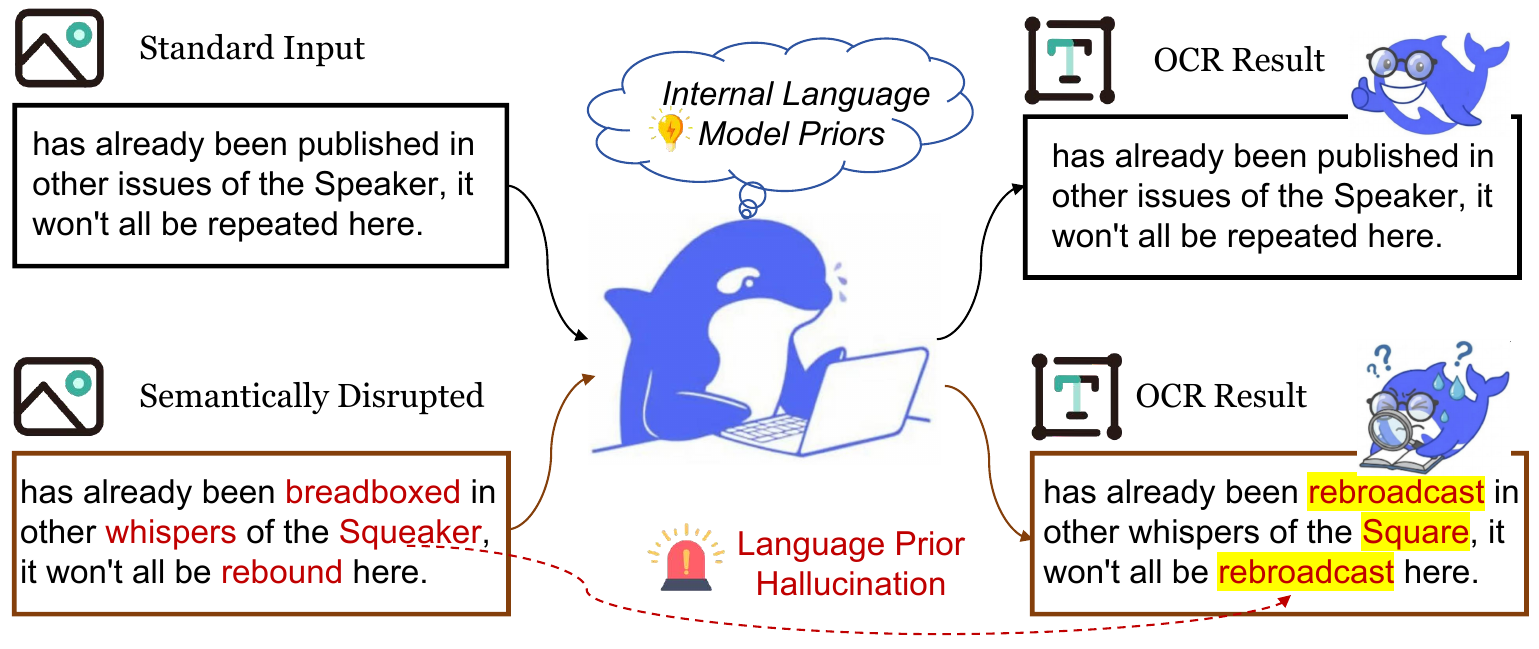}
    \caption{DeepSeek-OCR Model Over-reliance on Language Priors under Semantic Disruption.}
    \label{fig:prior_hallucination}
\end{figure}
Transformer-based Large Language Models (LLMs) face quadratic computational bottlenecks in long-context processing. 
Recent work DeepSeek-OCR \cite{deepseek-ocr} proposes an initial investigation into the feasibility of compressing long contexts via optical 2D mapping.
It encodes text into dense vision tokens using a vision encoder and decodes them back to text with a LLM-based text decoder.
It claims that a single page can be faithfully reconstructed from as few as 64–400 vision tokens, achieving 97\% OCR precision at less than 10× compression and approximately 60\% even at 20× compression on diverse layouts.
These results have been interpreted as evidence that visual modality can serve as an efficient compression medium for historical and long-form contexts in LLMs, opening new avenues for memory-efficient long-context processing.

However, we argue that these high OCR scores may not actually reflect genuine visual understanding, but rather an over-reliance on linguistic priors.
As shown in Figure \ref{fig:prior_hallucination}, DeepSeek-OCR employs a LLM as the text decoder, which inherently possesses strong statistical knowledge of language structure and common phrases.
When visual tokens are severely limited, the decoder may exploit these priors to "fill in the gaps," effectively guessing missing or ambiguous text based on context rather than visual evidence.
Thus, high scores may reflect linguistic crutches rather than visual merit, prompting our central question: Visual merit or linguistic crutch—which drives DeepSeek-OCR's performance?
To rigorously investigate this, we design a series of semantic disruption experiments with five key research questions:
\begin{itemize}
    \item \textbf{RQ1:How does sentence-level semantic disruption affect DeepSeek-OCR?}
    By introducing semantically absurd but visually similar sentence-level replacements, we probe global contextual priors' contribution.
    \item \textbf{RQ2:How does word-level semantic disruption affect DeepSeek-OCR?}
    Through intra-word letter swaps, full shuffles, and fully random character sequences devoid of any lexical or syntactic structure, we isolate local priors and measure pure visual contribution.
    \item \textbf{RQ3:How Does Linguistic Prior Dependence Manifest Across Different VLM and OCR Architectures?}
    We benchmark 13 OCR and VLM models with differing architectures on natural vs zero-prior random text to quantify the generality of prior exploitation.
    \item \textbf{RQ4:How does DeepSeek-OCR perform on QA and VQA tasks?}
    We further evaluate semantic fidelity for downstream document reasoning, including visual question answering (VQA) vs pure-text QA.
    \item \textbf{RQ5:What is the context length limit for optical compression?}
    Finally, we stress-test DeepSeek-OCR on real long-form narratives (up to 12,000 tokens) to identify practical scalability limits across its resolution modes.
\end{itemize}

Our experiments yield several critical findings:
1) Sentence-level semantic disruption causes substantial accuracy drops, especially under high compression (Tiny mode: -11.2\% avg; Small: -3.6\%; Base: -0.6\%), indicating global priors significantly aid reconstruction when visual tokens are scarce.
2) Word-level disruptions further degrade performance, with 10\% letter shuffles causing up to -11.3\% avg in Tiny mode, and fully random text collapsing accuracy to \textasciitilde20\%, confirming local priors also play a key role.
3) Benchmarking across 13 OCR/VLM models reveals all end-to-end architectures exhibit severe prior dependence, while traditional pipeline OCR methods show markedly higher robustness to semantic perturbations.
4) Downstream QA and VQA evaluations show that semantic integrity rapidly deteriorates under disruption, with VQA accuracy plummeting to near-random levels without linguistic cues.
5) Context stress tests demonstrate all DeepSeek-OCR modes fail between 8,000-10,500 tokens, suggesting optical compression may paradoxically exacerbate long-context bottlenecks rather than alleviate them.
We release all code, data, and scripts used in this study at \href{https://anonymous.4open.science/r/ACL26-84B5/Readme.md}{anonymous github}.

\begin{table*}[h]
    \centering
    \resizebox{\linewidth}{!}{ 
    \begin{tabular}{c|ccc|ccc|ccc}
        \toprule
        \multirow{3}{*}{Text Tokens} & \multicolumn{9}{c}{Precision (\%)} \\
        \cmidrule{2-10}
        & fox & fox & fox & text2png & text2png & text2png & distort & distort & distort \\
        & tiny & small & base & tiny & small & base & tiny & small & base \\
        \midrule
        600-700 & 95.93 & 98.30 & 98.51 & 98.64 & 99.00 & 99.56 & 96.23$^{\textcolor{red}{-2.41}}$ & 98.78$^{\textcolor{red}{-0.22}}$ & 99.60$^{\textcolor{green}{+0.04}}$ \\
        700-800 & 93.81 & 96.98 & 97.65 & 96.70 & 98.48 & 98.82 & 87.94$^{\textcolor{red}{-8.76}}$ & 94.08$^{\textcolor{red}{-4.40}}$ & 98.12$^{\textcolor{red}{-0.70}}$ \\
        800-900 & 91.91 & 96.65 & 97.74 & 94.46 & 97.49 & 98.75 & 88.46$^{\textcolor{red}{-6.00}}$ & 96.91$^{\textcolor{red}{-0.58}}$ & 99.20$^{\textcolor{green}{+0.45}}$ \\
        900-1000 & 84.19 & 96.68 & 98.80 & 87.73 & 96.94 & 98.94 & 70.06$^{\textcolor{red}{-17.67}}$ & 90.11$^{\textcolor{red}{-6.83}}$ & 98.20$^{\textcolor{red}{-0.74}}$ \\
        1000-1100 & 79.24 & 91.25 & 95.27 & 86.18 & 95.24 & 96.67 & 74.54$^{\textcolor{red}{-11.64}}$ & 92.05$^{\textcolor{red}{-3.19}}$ & 96.32$^{\textcolor{red}{-0.35}}$ \\
        1100-1200 & 74.34 & 89.21 & 93.67 & 80.72 & 94.21 & 95.46 & 57.76$^{\textcolor{red}{-22.96}}$ & 87.11$^{\textcolor{red}{-7.10}}$ & 95.03$^{\textcolor{red}{-0.43}}$ \\
        1200-1300 & 58.73 & 86.44 & 89.65 & 73.97 & 91.85 & 93.39 & 54.26$^{\textcolor{red}{-19.71}}$ & 90.23$^{\textcolor{red}{-1.62}}$ & 97.62$^{\textcolor{green}{+4.23}}$ \\
        1300-1400 & 69.34 & 90.98 & 96.05 & 64.22 & 87.61 & 98.16 & 50.49$^{\textcolor{red}{-13.73}}$ & 81.22$^{\textcolor{red}{-6.39}}$ & 96.05$^{\textcolor{red}{-2.11}}$ \\
        1400-1500 & 76.70 & 96.03 & 99.47 & 64.48 & 90.86 & 98.68 & 1.94$^{\textcolor{red}{-62.54}}$ & 73.33$^{\textcolor{red}{-17.53}}$ & 96.75$^{\textcolor{red}{-1.93}}$ \\
        1500-1600 & 34.41 & 76.01 & 92.23 & 53.59 & 84.41 & 95.83 & 46.75$^{\textcolor{red}{-6.84}}$ & 80.46$^{\textcolor{red}{-3.95}}$ & 98.18$^{\textcolor{green}{+2.35}}$ \\
        1600-1700 & 58.14 & 86.18 & 94.02 & 64.50 & 87.01 & 97.13 & 44.12$^{\textcolor{red}{-20.38}}$ & 79.54$^{\textcolor{red}{-7.47}}$ & 96.97$^{\textcolor{red}{-0.16}}$ \\
        1700-1800 & 34.43 & 76.29 & 95.78 & 60.00 & 74.34 & 94.29 & 25.93$^{\textcolor{red}{-34.07}}$ & 20.00$^{\textcolor{red}{-54.34}}$ & 43.14$^{\textcolor{red}{-51.15}}$ \\
        2400-2500 & 0.43 & 37.24 & 73.60 & 13.66 & 13.61 & 84.00 & 40.59$^{\textcolor{green}{+26.93}}$ & 63.92$^{\textcolor{green}{+50.31}}$ & 81.05$^{\textcolor{red}{-2.95}}$ \\
        \midrule
        Average & 83.88 & 93.89 & 96.62 & 88.00 & 95.23 & 97.93 & 76.75$^{\textcolor{red}{-11.25}}$ & 91.56$^{\textcolor{red}{-3.67}}$ & 97.31$^{\textcolor{red}{-0.62}}$ \\
        \bottomrule
    \end{tabular}
    }
    \caption{Performance of DeepSeek-OCR Under Sentence-Level Semantic Disruption.}
    \vspace{-1em}
    \label{tab:ocr_token_length}
\end{table*}

\section{RQ1: How Does Sentence-Level Semantic Disruption Affect DeepSeek-OCR?}
\label{rq1}

\subsection{Experimental Setup}
We base our evaluation on the Fox benchmark \cite{fox}, comprising 112 English document pages with ground-truth token lengths ranging from 600 to 2500.
As a clean baseline, we render the ground-truth text into images (text2png).
For disruption, we apply targeted replacements guided by a controlled distortion process: key nouns, verbs, and phrases are substituted with absurd alternatives mimicking English patterns (e.g., "butterfly" → "breadflutter"), preserving character shapes and layout while eliminating meaningful context.
The distorted text is also rendered into images (distort) using the same pipeline.
Both text2png and distort sets are evaluated on DeepSeek-OCR in Tiny, Small, and Base modes. 
We also report results on the original Fox images for reference, and performance is measured by OCR precision.

\subsection{Results and Analysis}
Table \ref{tab:ocr_token_length} reports precision across text token length bins.
We can find that introducing sentence-level semantic disruption substantially degrades accuracy, particularly under higher compression modes.
In Tiny mode, distort reduces average precision to 76.7\% (-11.2\%), while Small and Base modes see smaller but still significant drops to 91.5\% (-3.6\%) and 97.3\% (-0.6\%), respectively.

These patterns demonstrate that sentence-level semantic priors serve as a significant linguistic crutch when visual tokens are limited. 
These significant drops in Tiny mode indicate that when visual tokens are extremely limited, the decoder heavily relies on global linguistic context to reconstruct plausible text.
In contrast, ample vision tokens (Base) enable near-perfect recovery regardless of semantic validity, indicating that sufficient visual resolution reduces dependence on higher-order linguistic priors.
Overall, sentence-level semantic disruption reveals a clear trade-off: high reported OCR accuracy under compression is partly an illusion, as it's sustained by the decoder’s ability to exploit global linguistic context rather than genuine visual understanding.

\subsection{Case Study}
Figure \ref{fig:prior_hallucination} illustrates a concrete example of prior-induced hallucination under sentence-level disruption.
The original text is: "has already been published in other issues of the Speaker, it won't all be repeated here.", 
and we replace "published" with "breadboxed", "issues" with "whispers", "Speaker" with "Squeaker", and "repeated" with "rebound" to create the disrupted text: "has already been breadboxed in other whispers of the Squeaker, it won't all be rebound here."
For the original text, DeepSeek-OCR produces a minor contextual shift ("Special Issue" instead of "Speaker issues"), likely corrected by priors. 
For the disrupted text, we can find that when faced with the visually clear but semantically absurd word "Squeaker", DeepSeek-OCR fails to transcribe the visual input faithfully. Instead, it hallucinates the word "Square" and attempts to "correct" the non-existent word "rebound" into "rebroadcast".
This behavior confirms that the model prioritizes linguistic probability over visual evidence.

\begin{table*}[t]
    \centering
    \adjustbox{width=0.85\textwidth}{
    \begin{tabular}{c|ccc|ccc}
        \toprule
        \multirow{2}{*}{Text Tokens} & \multicolumn{3}{c}{Swap (\%)} & \multicolumn{3}{c}{Shuffle (\%)} \\
        \cmidrule(lr){2-4} \cmidrule(lr){5-7}
        & tiny & small & base & tiny & small & base \\
        \midrule
        600-700 & 94.19$^{\textcolor{red}{-4.45}}$ & 95.21$^{\textcolor{red}{-3.79}}$ & 96.55$^{\textcolor{red}{-3.01}}$ & 87.37$^{\textcolor{red}{-11.27}}$ & 91.83$^{\textcolor{red}{-7.17}}$ & 96.72$^{\textcolor{red}{-2.84}}$ \\
        700-800 & 92.02$^{\textcolor{red}{-4.68}}$ & 94.94$^{\textcolor{red}{-3.54}}$ & 95.94$^{\textcolor{red}{-2.88}}$ & 85.81$^{\textcolor{red}{-10.89}}$ & 88.53$^{\textcolor{red}{-9.95}}$ & 94.16$^{\textcolor{red}{-4.66}}$ \\
        800-900 & 90.82$^{\textcolor{red}{-3.64}}$ & 93.92$^{\textcolor{red}{-3.57}}$ & 96.16$^{\textcolor{red}{-2.59}}$ & 83.17$^{\textcolor{red}{-11.29}}$ & 89.55$^{\textcolor{red}{-7.94}}$ & 95.56$^{\textcolor{red}{-3.19}}$ \\
        900-1000 & 84.21$^{\textcolor{red}{-3.52}}$ & 92.31$^{\textcolor{red}{-4.63}}$ & 95.72$^{\textcolor{red}{-3.22}}$ & 77.48$^{\textcolor{red}{-10.25}}$ & 87.32$^{\textcolor{red}{-9.62}}$ & 93.02$^{\textcolor{red}{-5.92}}$ \\
        1000-1100 & 81.42$^{\textcolor{red}{-4.76}}$ & 91.73$^{\textcolor{red}{-3.51}}$ & 94.49$^{\textcolor{red}{-2.18}}$ & 73.93$^{\textcolor{red}{-12.25}}$ & 85.40$^{\textcolor{red}{-9.84}}$ & 92.63$^{\textcolor{red}{-4.04}}$ \\
        1100-1200 & 78.13$^{\textcolor{red}{-2.59}}$ & 89.98$^{\textcolor{red}{-4.23}}$ & 94.08$^{\textcolor{red}{-1.38}}$ & 72.87$^{\textcolor{red}{-7.85}}$ & 84.76$^{\textcolor{red}{-9.45}}$ & 93.26$^{\textcolor{red}{-2.20}}$ \\
        1200-1300 & 69.70$^{\textcolor{red}{-4.27}}$ & 87.08$^{\textcolor{red}{-4.77}}$ & 94.56$^{\textcolor{green}{+1.17}}$ & 62.56$^{\textcolor{red}{-11.41}}$ & 83.05$^{\textcolor{red}{-8.80}}$ & 92.85$^{\textcolor{red}{-0.54}}$ \\
        1300-1400 & 61.07$^{\textcolor{red}{-3.15}}$ & 85.31$^{\textcolor{red}{-2.30}}$ & 95.42$^{\textcolor{red}{-2.74}}$ & 56.58$^{\textcolor{red}{-7.64}}$ & 78.39$^{\textcolor{red}{-9.22}}$ & 92.84$^{\textcolor{red}{-5.32}}$ \\
        1400-1500 & 67.63$^{\textcolor{green}{+3.15}}$ & 91.58$^{\textcolor{green}{+0.72}}$ & 95.42$^{\textcolor{red}{-3.26}}$ & 25.23$^{\textcolor{red}{-39.25}}$ & 82.66$^{\textcolor{red}{-8.20}}$ & 90.65$^{\textcolor{red}{-8.03}}$ \\
        1500-1600 & 37.38$^{\textcolor{red}{-16.21}}$ & 78.82$^{\textcolor{red}{-5.59}}$ & 91.92$^{\textcolor{red}{-3.91}}$ & 40.32$^{\textcolor{red}{-13.27}}$ & 71.56$^{\textcolor{red}{-12.85}}$ & 92.69$^{\textcolor{red}{-3.14}}$ \\
        1600-1700 & 55.71$^{\textcolor{red}{-8.79}}$ & 83.86$^{\textcolor{red}{-3.15}}$ & 95.51$^{\textcolor{red}{-1.62}}$ & 43.50$^{\textcolor{red}{-21.00}}$ & 78.12$^{\textcolor{red}{-8.89}}$ & 93.42$^{\textcolor{red}{-3.71}}$ \\
        1700-1800 & 25.45$^{\textcolor{red}{-34.55}}$ & 21.05$^{\textcolor{red}{-53.29}}$ & 89.66$^{\textcolor{red}{-4.63}}$ & 37.93$^{\textcolor{red}{-22.07}}$ & 61.54$^{\textcolor{red}{-12.80}}$ & 84.69$^{\textcolor{red}{-9.60}}$ \\
        2400-2500 & 31.11$^{\textcolor{green}{+17.45}}$ & 60.50$^{\textcolor{green}{+46.89}}$ & 88.18$^{\textcolor{green}{+4.18}}$ & 1.94$^{\textcolor{red}{-11.72}}$ & 56.54$^{\textcolor{green}{+42.93}}$ & 87.76$^{\textcolor{green}{+3.76}}$ \\
        \midrule
        Average & 83.69$^{\textcolor{red}{-4.31}}$ & 91.51$^{\textcolor{red}{-3.72}}$ & 95.45$^{\textcolor{red}{-2.48}}$ & 81.83$^{\textcolor{red}{-6.17}}$ & 91.24$^{\textcolor{red}{-3.99}}$ & 95.84$^{\textcolor{red}{-2.09}}$ \\
        \bottomrule
    \end{tabular}
    }
    \caption{Performance of DeepSeek-OCR under 5\% word-level semantic corruption.}
    \label{tab:swap_shuffle_5}
\end{table*}

\begin{table*}[t]
    \centering
        \adjustbox{width=0.85\textwidth}{
    \begin{tabular}{c|ccc|ccc}
        \toprule
        \multirow{2}{*}{Text Tokens} & \multicolumn{3}{c}{Swap (\%)} & \multicolumn{3}{c}{Shuffle (\%)} \\
        \cmidrule(lr){2-4} \cmidrule(lr){5-7}
        & tiny & small & base & tiny & small & base \\
        \midrule
        600-700 & 90.14$^{\textcolor{red}{-8.50}}$ & 92.60$^{\textcolor{red}{-6.40}}$ & 95.51$^{\textcolor{red}{-4.05}}$ & 87.37$^{\textcolor{red}{-11.27}}$ & 91.83$^{\textcolor{red}{-7.17}}$ & 96.72$^{\textcolor{red}{-2.84}}$ \\
        700-800 & 87.82$^{\textcolor{red}{-8.88}}$ & 89.82$^{\textcolor{red}{-8.66}}$ & 94.03$^{\textcolor{red}{-4.79}}$ & 85.81$^{\textcolor{red}{-10.89}}$ & 88.53$^{\textcolor{red}{-9.95}}$ & 94.16$^{\textcolor{red}{-4.66}}$ \\
        800-900 & 86.29$^{\textcolor{red}{-8.17}}$ & 90.84$^{\textcolor{red}{-6.65}}$ & 94.45$^{\textcolor{red}{-4.30}}$ & 83.17$^{\textcolor{red}{-11.29}}$ & 89.55$^{\textcolor{red}{-7.94}}$ & 95.56$^{\textcolor{red}{-3.19}}$ \\
        900-1000 & 73.06$^{\textcolor{red}{-14.67}}$ & 88.91$^{\textcolor{red}{-8.03}}$ & 93.90$^{\textcolor{red}{-5.04}}$ & 77.48$^{\textcolor{red}{-10.25}}$ & 87.32$^{\textcolor{red}{-9.62}}$ & 93.02$^{\textcolor{red}{-5.92}}$ \\
        1000-1100 & 77.16$^{\textcolor{red}{-9.02}}$ & 88.09$^{\textcolor{red}{-7.15}}$ & 92.07$^{\textcolor{red}{-4.60}}$ & 73.93$^{\textcolor{red}{-12.25}}$ & 85.40$^{\textcolor{red}{-9.84}}$ & 92.63$^{\textcolor{red}{-4.04}}$ \\
        1100-1200 & 74.62$^{\textcolor{red}{-6.10}}$ & 87.09$^{\textcolor{red}{-7.12}}$ & 92.93$^{\textcolor{red}{-2.53}}$ & 72.87$^{\textcolor{red}{-7.85}}$ & 84.76$^{\textcolor{red}{-9.45}}$ & 93.26$^{\textcolor{red}{-2.20}}$ \\
        1200-1300 & 48.84$^{\textcolor{red}{-25.13}}$ & 85.85$^{\textcolor{red}{-6.00}}$ & 94.93$^{\textcolor{green}{+1.54}}$ & 62.56$^{\textcolor{red}{-11.41}}$ & 83.05$^{\textcolor{red}{-8.80}}$ & 92.85$^{\textcolor{red}{-0.54}}$ \\
        1300-1400 & 58.17$^{\textcolor{red}{-6.05}}$ & 80.99$^{\textcolor{red}{-6.62}}$ & 93.28$^{\textcolor{red}{-4.88}}$ & 56.58$^{\textcolor{red}{-7.64}}$ & 78.39$^{\textcolor{red}{-9.22}}$ & 92.84$^{\textcolor{red}{-5.32}}$ \\
        1400-1500 & 64.94$^{\textcolor{green}{+0.46}}$ & 86.89$^{\textcolor{red}{-3.97}}$ & 93.87$^{\textcolor{red}{-4.81}}$ & 25.23$^{\textcolor{red}{-39.25}}$ & 82.66$^{\textcolor{red}{-8.20}}$ & 90.65$^{\textcolor{red}{-8.03}}$ \\
        1500-1600 & 51.10$^{\textcolor{red}{-2.49}}$ & 73.37$^{\textcolor{red}{-11.04}}$ & 91.61$^{\textcolor{red}{-4.22}}$ & 40.32$^{\textcolor{red}{-13.27}}$ & 71.56$^{\textcolor{red}{-12.85}}$ & 92.69$^{\textcolor{red}{-3.14}}$ \\
        1600-1700 & 41.52$^{\textcolor{red}{-22.98}}$ & 79.64$^{\textcolor{red}{-7.37}}$ & 92.27$^{\textcolor{red}{-4.86}}$ & 43.50$^{\textcolor{red}{-21.00}}$ & 78.12$^{\textcolor{red}{-8.89}}$ & 93.42$^{\textcolor{red}{-3.71}}$ \\
        1700-1800 & 31.87$^{\textcolor{red}{-28.13}}$ & 59.15$^{\textcolor{red}{-15.19}}$ & 86.09$^{\textcolor{red}{-8.20}}$ & 37.93$^{\textcolor{red}{-22.07}}$ & 61.54$^{\textcolor{red}{-12.80}}$ & 84.69$^{\textcolor{red}{-9.60}}$ \\
        2400-2500 & 5.30$^{\textcolor{red}{-8.36}}$ & 46.86$^{\textcolor{green}{+33.25}}$ & 89.19$^{\textcolor{green}{+5.19}}$ & 1.94$^{\textcolor{red}{-11.72}}$ & 56.54$^{\textcolor{green}{+42.93}}$ & 87.76$^{\textcolor{green}{+3.76}}$ \\
        \midrule
        Average & 78.11$^{\textcolor{red}{-9.89}}$ & 88.07$^{\textcolor{red}{-7.16}}$ & 93.75$^{\textcolor{red}{-4.18}}$ & 76.70$^{\textcolor{red}{-11.30}}$ & 86.55$^{\textcolor{red}{-8.68}}$ & 93.99$^{\textcolor{red}{-3.94}}$ \\
        \bottomrule
    \end{tabular}
    }
    \caption{Performance of DeepSeek-OCR under 10\% word-level semantic corruption.}
    \vspace{-1em}
    \label{tab:swap_shuffle_10}
\end{table*}

\begin{table}[t]
    \centering
    \adjustbox{width=0.35\textwidth}{
    \begin{tabular}{cccc}
        \toprule
        \multirow{2}{*}{Text Tokens} & \multicolumn{3}{c}{Precision (\%)} \\
        \cmidrule(lr){2-4}
        & tiny & small & base \\
        \midrule
        600-700 & 26.81 & 45.37 & 63.54 \\
        700-800 & 24.29 & 45.25 & 61.93 \\
        800-900 & 22.77 & 46.53 & 63.47 \\
        900-1000 & 23.44 & 42.44 & 60.94 \\
        1000-1100 & 15.44 & 43.15 & 61.07 \\
        1100-1200 & 15.21 & 40.35 & 61.60 \\
        1200-1300 & 11.20 & 33.48 & 60.18 \\
        1300-1400 & 8.40 & 33.02 & 62.36 \\
        1400-1500 & 5.23 & 25.87 & 58.28 \\
        1500-1600 & 2.48 & 23.21 & 57.83 \\
        1600-1700 & 4.43 & 24.00 & 56.05 \\
        1700-1800 & 2.86 & 18.91 & 56.64 \\
        2400-2500 & 0.00 & 6.97 & 45.60 \\
        \midrule
        Average & 19.84 & 42.12 & 61.70 \\
        \bottomrule
    \end{tabular}
    }
    \caption{OCR performance with unsemantic samples.}
    \label{tab:random}
\end{table}

\section{RQ2: How Does Word-Level Semantic Disruption Affect DeepSeek-OCR?}
Having established the role of sentence-level priors, we now turn to finer-grained disruptions at the word level. 

\subsection{Experimental Setup}
We continue using the Fox benchmark as the baseline with three word-level perturbation strategies:
\begin{itemize}
    \item \textbf{Swap}: Randomly select 5\% or 10\% of words and swap two letters within each selected word, 
    creating minor spelling distortions that preserve most word structure but introduce errors repairable by linguistic priors.
    \item \textbf{Shuffle}: Randomly select 5\% or 10\% of words and fully shuffle the letters within each selected word, destroying internal word structure while keeping character distributions similar.
    \item \textbf{Zero-Prior Random Text}: Generate entirely new “words” (2–10 random letters, mixed case) to form documents with identical length distribution of the original Fox instances, but devoid of any lexical or syntactic structure.
\end{itemize}

\subsection{Results and Analysis}
The performance degradation across these settings (Table \ref{tab:swap_shuffle_5},\ref{tab:swap_shuffle_10},\ref{tab:random}) exposes a heavy dependency on lexical priors.
First, we can find that DeepSeek-OCR is sensitive to letter order, even minor disruptions cause disproportionate failures in high-compression modes. At 10\% disruption, Tiny mode precision drops by 9.89\% (Swap) and 11.30\% (Shuffle) on average.
Notably, the sharper decline in Shuffle confirms that the model relies on standard character ordering (n-grams) to decode text; when this order is disrupted, the "visual" recovery fails. 
Second, the most compelling evidence comes from the Zero-Prior experiment. When denied meaningful words, DeepSeek-OCR's performance suffers a catastrophic collapse, precision in Tiny mode plummets to a mere 19.84\%. 
Third, the huge gap between high scores on natural text(\textasciitilde90\%) vs near-random text (\textasciitilde20\%) conclusively proves that most of its reported "accuracy" in compressed modes is derived from linguistic hallucination, not visual recognition.



\begin{table}[ht]
\centering
\footnotesize
\begin{tabularx}{\columnwidth}{@{} l X @{}}
    \toprule
    \textbf{Scenario} & \textbf{Comparison (Input $\rightarrow$ OCR Result)} \\ \midrule
    \multirow{2}{*}{\textbf{Swap}} & \textbf{Input:} ...has \underline{ayreadl} ... \\
    & \textbf{Output:} ...has \textbf{already} ... \\
    \midrule
    \multirow{2}{*}{\textbf{Shuffle}} & \textbf{Input:} ...it won't all be \underline{eepetadr} here. \\
    & \textbf{Output:} ...I won't be all \textbf{expected} here. \\
    \midrule
    \multirow{2}{*}{\textbf{Zero-Prior}} & \textbf{Input:} EYuoV qUtjpy pWxZCks vUQnwh K qCuYCXmor \\
    & \textbf{Output:} E'vuol u'qtippy piwZckus u'Quwnh kq'CuYcKorom \\
    \bottomrule
\end{tabularx}
\caption{Case of Model Behavior under Different Word-Level Disruptions}
\vspace{-1em}
\label{tab:case2}
\end{table}

\subsection{Case Study}
As shown in Table \ref{tab:case2}, we continue with the example in RQ1 to illustrate word-level disruptions.
In the Swap scenario ("ayreadl" instead of "already"), the model acts as an auto-corrector, outputting the correct English word "already" despite the visual mismatch. This shows reliance on lexical priors to fix minor errors.
In the Shuffle scenario ("eepetadr" instead of "repeated"), the model again leverages context to produce "expected", a plausible English word, rather than attempting to decode the nonsensical input.
Finally, in the Zero-Prior Random Text scenario, the model fails entirely. It attempts to force-fit the random visual patterns into quasi-syllabic structures, resulting in output that is neither the ground truth nor a valid word, but a manifestation of the decoder struggling without priors.
\begin{table*}[htbp]
\centering
\resizebox{\linewidth}{!}{ 
\begin{tabular}{lccccccccccc|cc}
\toprule
\multirow{4}{*}{Text Tokens} & \multicolumn{13}{c}{Precision (\%)} \\
\cmidrule(lr){2-14}
& \multicolumn{11}{c}{End-to-End} & \multicolumn{2}{c}{Pipeline} \\
\cmidrule(lr){2-12} \cmidrule(lr){13-14}
& \multicolumn{2}{c}{DeepSeek-OCR} & dots.ocr & \multicolumn{2}{c}{Qwen2.5vl} & GOT-OCR & \multicolumn{2}{c}{MonkeyOCR} & SmolDocling & Nougat & HunyuanOCR & MinerU & PaddleOCR-v5 \\
\cmidrule(lr){2-3} \cmidrule(lr){5-6} \cmidrule(lr){8-9}
& tiny & small & 7B & 7B & 72B & 0.58B & 1.2B & 3B & 0.125B & 0.35B & 1B & 1.2B & 0.07B \\
\midrule
600-700    & 98.64 & 99.00 & 98.99 & 99.52 & 99.73 & 99.56 & 98.48 & 99.46 & 98.90 & 99.02 & 99.50 & 10.20 & 97.82 \\
700-800    & 96.70 & 98.48 & 97.12 & 98.02 & 98.33 & 98.16 & 97.34 & 97.53 & 90.34 & 97.17 & 98.23 & 9.00  & 95.73 \\
800-900    & 94.46 & 97.49 & 97.51 & 98.23 & 98.76 & 98.75 & 97.00 & 97.28 & 93.60 & 96.85 & 97.53 & 9.12  & 95.39 \\
900-1000   & 87.73 & 96.94 & 98.12 & 98.98 & 99.04 & 99.11 & 97.37 & 97.90 & 98.17 & 98.09 & 98.61 & 11.70 & 96.17 \\
1000-1100  & 86.18 & 95.24 & 97.65 & 96.97 & 98.20 & 97.67 & 94.42 & 96.97 & 96.84 & 96.49 & 97.57 & 7.67  & 93.68 \\
1100-1200  & 80.72 & 94.21 & 95.98 & 96.73 & 98.21 & 96.30 & 93.97 & 95.39 & 93.42 & 93.20 & 96.09 & 7.15  & 91.18 \\
1200-1300  & 73.97 & 91.85 & 97.71 & 98.29 & 99.61 & 98.08 & 93.74 & 95.27 & 95.33 & 95.43 & 97.79 & 8.65  & 88.46 \\
1300-1400  & 64.22 & 87.61 & 97.44 & 98.05 & 99.68 & 98.22 & 97.26 & 97.97 & 97.52 & 98.39 & 97.32 & 9.61  & 94.63 \\
1400-1500  & 64.48 & 90.86 & 99.47 & 1.00  & 99.74 & 98.15 & 99.47 & 99.21 & 98.95 & 97.61 & 99.74 & 17.48 & 95.23 \\
1500-1600  & 53.59 & 84.41 & 92.95 & 97.85 & 97.84 & 95.87 & 93.93 & 95.23 & 91.28 & 92.62 & 95.68 & 1.87  & 90.23 \\
1600-1700  & 64.50 & 87.01 & 95.56 & 97.29 & 97.63 & 95.42 & 96.66 & 96.51 & 95.88 & 97.07 & 95.80 & 6.84  & 84.13 \\
1700-1800  & 60.00 & 74.34 & 97.32 & 96.49 & 96.49 & 90.43 & 100.00& 100.00& 94.77 & 92.87 & 96.98 & 0.00  & 93.61 \\
2400-2500  & 13.66 & 13.61 & 98.61 & 99.43 & 100.00& 78.47 & 92.71 & 100.00& 94.92 & 94.27 & 99.43 & 0.00  & 70.37 \\
\midrule
Average    & 88.00 & 95.23 & 97.40 & 98.10 & 98.69 & 98.00 & 96.60 & 97.37 & 94.30 & 96.82 & 97.83 & 8.94  & 94.44 \\
\bottomrule
\end{tabular}
}
\caption{Comparison with other VLM and OCR models.}
\label{tab:from_text_other_models}
\end{table*}

\begin{table*}[t]
\centering
\resizebox{\linewidth}{!}{ 
\begin{tabular}{lccccccccccc|cc} 
\toprule
\multirow{4}{*}{Text Tokens} & \multicolumn{13}{c}{Precision (\%)} \\ 
\cmidrule(lr){2-14} 
& \multicolumn{11}{c}{End-to-End} & \multicolumn{2}{c}{Pipeline} \\
\cmidrule(lr){2-12} \cmidrule(lr){13-14}
& \multicolumn{2}{c}{DeepSeek-OCR} & dots.ocr & \multicolumn{2}{c}{Qwen2.5vl} & GOT-OCR & \multicolumn{2}{c}{MonkeyOCR} & SmolDocling & Nougat & HunyuanOCR & MinerU & PaddleOCR-v5 \\
\cmidrule(lr){2-3} \cmidrule(lr){5-6} \cmidrule(lr){8-9} 
& tiny & small & 7B & 7B & 72B & 0.58B & 1.2B & 3B & 0.125B & 0.35B & 1B & 1.2B & 0.07B \\
\midrule
600-700   & 26.81 & 45.37 & 46.80 & 51.20 & 58.61 & 70.13 & 71.77 & 80.39 & 66.45 & 45.16 & 47.80 & 4.61 & 88.94 \\
700-800   & 24.29 & 45.25 & 47.77 & 50.49 & 56.14 & 59.63 & 71.63 & 79.64 & 61.10 & 36.50 & 40.76 & 3.01 & 89.93 \\
800-900   & 22.77 & 46.53 & 48.83 & 50.54 & 58.11 & 63.43 & 72.53 & 80.30 & 58.28 & 40.50 & 35.98 & 4.39 & 89.70 \\
900-1000  & 23.44 & 42.44 & 43.97 & 44.11 & 54.55 & 57.50 & 71.17 & 79.34 & 54.29 & 42.36 & 35.95 & 4.25 & 89.59 \\
1000-1100 & 15.44 & 43.15 & 46.69 & 38.83 & 59.35 & 55.23 & 72.07 & 80.86 & 45.34 & 38.58 & 38.44 & 4.09 & 89.00 \\
1100-1200 & 15.21 & 40.35 & 45.65 & 49.88 & 58.25 & 55.87 & 71.71 & 80.68 & 54.21 & 39.59 & 36.45 & 4.46 & 89.66 \\
1200-1300 & 11.20 & 33.48 & 43.98 & 38.00 & 55.57 & 49.34 & 70.16 & 79.44 & 59.58 & 15.51 & 38.67 & 2.85 & 89.43 \\
1300-1400 & 8.40  & 33.02 & 40.26 & 37.78 & 54.10 & 47.15 & 70.91 & 79.97 & 58.03 & 39.06 & 34.21 & 0.86 & 89.59 \\
1400-1500 & 5.23  & 25.87 & 44.16 & 43.45 & 53.61 & 45.33 & 70.84 & 80.38 & 56.95 & 36.57 & 34.29 & 4.75 & 99.96 \\
1500-1600 & 2.48  & 23.21 & 40.95 & 48.66 & 53.30 & 40.58 & 68.64 & 79.56 & 49.05 & 21.83 & 31.51 & 1.43 & 88.94 \\
1600-1700 & 4.43  & 24.00 & 39.35 & 24.22 & 51.74 & 35.60 & 67.74 & 76.64 & 46.03 & 37.74 & 36.12 & 0.00 & 87.55 \\
1700-1800 & 2.86  & 18.91 & 38.88 & 41.89 & 56.18 & 29.68 & 65.39 & 77.78 & 54.39 & 0.00  & 30.45 & 0.00 & 86.98 \\
2400-2500 & 0.00  & 6.97  & 32.89 & 44.87 & 50.16 & 10.69 & 66.72 & 80.13 & 38.57 & 36.06 & 31.11 & 4.17 & 88.98 \\
\midrule
Average   & 19.84 & 42.12 & 46.31 & 46.86 & 56.78 & 57.71 & 71.55 & 79.97 & 56.77 & 37.94 & 38.04 & 3.67 & 89.53\\
\bottomrule
\end{tabular}
}
\caption{Performance with completely unsemantic samples on VLM and OCR models.}
\vspace{-1em}
\label{tab:random_other_models}
\end{table*}

\section{RQ3: How Does Linguistic Prior Dependence Manifest Across Different VLM and OCR Architectures?}
To determine whether linguistic prior dependence is a unique flaw of DeepSeek-OCR or a broader phenomenon in other VLMs and OCR systems, we conduct a comparative analysis across diverse architectures.
\subsection{Experimental Setup}
We compare DeepSeek-OCR (Tiny/Small modes) against a diverse set of 11 additional models spanning 125M-72B parameters on the clean natural text and zero-prior random text.
The zero-prior random text serves as a "truth serum" for OCR systems, forcing them to rely solely on visual recognition capabilities.
\subsection{Results and Analysis}
The results in Table \ref{tab:from_text_other_models},\ref{tab:random_other_models} reveal a distinct architecture disparity.
On the natural text, end-to-end models achieve impressive precision (\textasciitilde97-98\%), comparable to DeepSeek-OCR.
However, when handling with zero-prior random text, end-to-end models suffer catastrophic performance collapses, dropping 40-60\% in precision.
DeepSeek-OCR (Tiny) suffers a massive 68.16\% drop in precision. Similarly, peer end-to-end models exhibit catastrophic declines: HunyuanOCR (-59.79\%), Nougat (-58.88\%), and Qwen2.5-VL 7B (-51.24\%).
This corroborates that end-to-end architectures heavily rely on linguistic priors to compensate for visual recognition shortcomings.
In contrast, traditional pipeline OCR model PaddleOCR-v5 \cite{paddleocr} demonstrates remarkable resilience, with only a minor 4.9\% precision drop to 89.53\%.
Unlike end-to-end models that predict text directly from images, pipeline OCR systems separate visual recognition from linguistic decoding, allowing them to maintain performance even when linguistic priors are absent.
Notably, MinerU \cite{mineru} is also a pipeline system, but it performs poorly (under 10\%) because its detection model misidentifies the entire image as a single bounding box, leading to ineffective OCR processing.




\subsection{Case Study}
As shown in Figure \ref{fig:case3}, we present an example of OCR results on natural text and random text across end-to-end OCR model (DeepSeek-OCR Small), VLM (Qwen2.5-VL 72B), and traditional pipeline OCR model (PaddleOCR-v5).
For natural text, all three models perform perfectly, achieving 100\% precision.
However, on random text, both DeepSeek-OCR Small and Qwen2.5-VL 72B can hardly recognize correct words, struggle to force-fit the random visual patterns into known tokens.
In contrast, PaddleOCR-v5 maintains a high precision, correctly identifying most of the unsemantic words, demonstrating its robustness without reliance on linguistic priors.
This comparative analysis confirms that while end-to-end optical compression models (like DeepSeek-OCR) excel in token efficiency, they sacrifice the intrinsic visual robustness that is inherent to traditional pipeline systems. 
They do not merely "read" text; they reconstruct it through a linguistic lens, which becomes a liability when the text is unstructured or nonsensical.



\section{RQ4: How Does DeepSeek-OCR Perform on QA and VQA Tasks?}
High OCR accuracy does not guarantee preservation of semantic content necessary for downstream reasoning. 
To evaluate this, we compare performance on document QA and VQA.

\subsection{Experimental Setup}
We extend the Fox benchmark by annotating each document pages with three fact-based question-answer pairs. 
VQA includes strong multimodal baselines: Qwen2.5VL-3B/7B, Qwen3VL-4B/8B \cite{qwen3vl}, and MiniCPM-V 4.5 \cite{minicpm}.
QA Baselines include Qwen2.5-3B, Qwen3-4B, and Llama3.2-3B (all \textasciitilde3–4B scale, similar to DeepSeek-OCR’s activated parameters). 

\subsection{Results and Analysis}
\begin{figure}[t]
    \centering
    \includegraphics[width=0.47\textwidth]{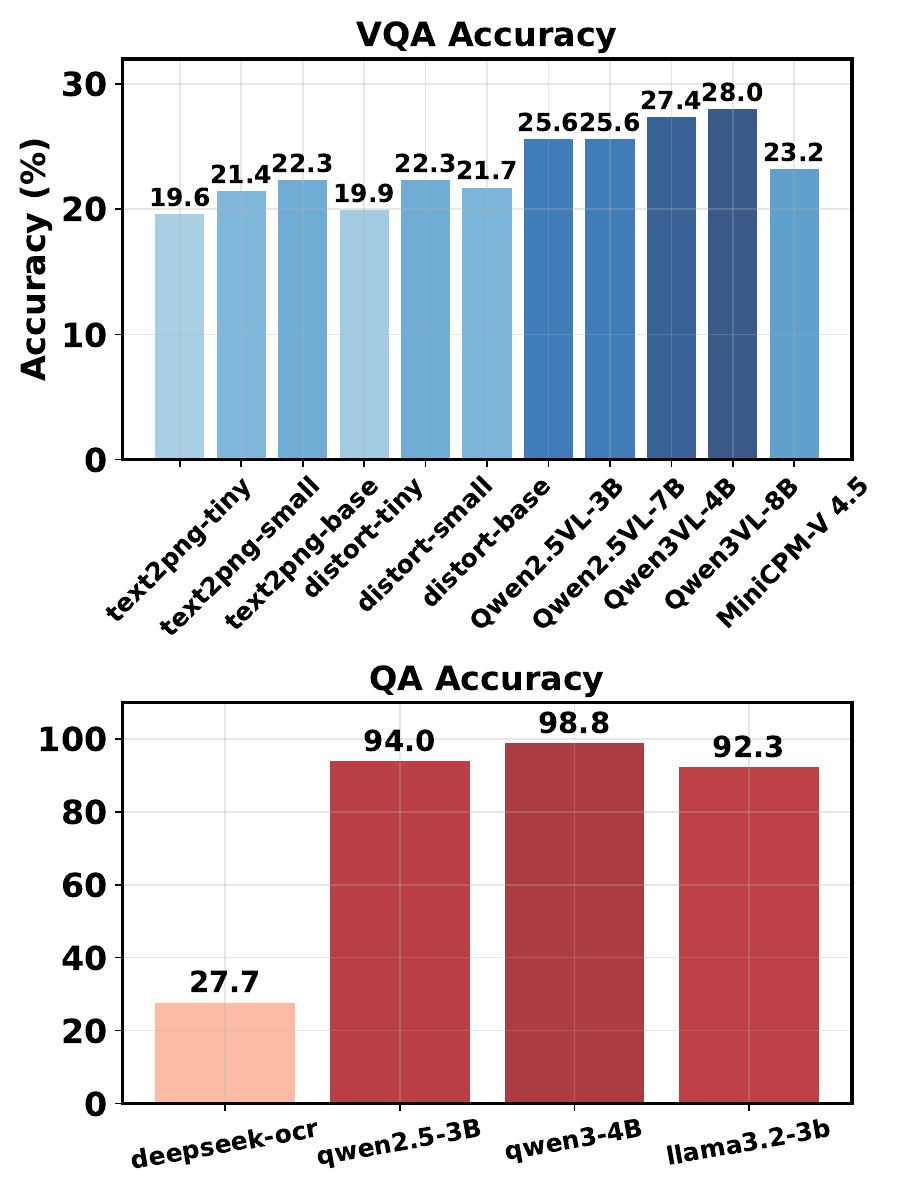}
    \caption{VQA and QA Performance.}
    \vspace{-1em}
    \label{fig:vqa_qa}
\end{figure}

The performance gap illustrated in Figure \ref{fig:vqa_qa} is striking and reveals a fundamental limitation of optical compression for preserving semantic content.
We can find there's a reasoning collapse in VQA: while DeepSeek-OCR claims high OCR precision, its performance on VQA is near random chance (\textasciitilde20\% accuracy for four-option questions). This indicates that the visual representations, while sufficient to trigger the decoder's linguistic priors for text reconstruction, fail to capture the deeper semantic relationships needed for logical reasoning.

In sharp contrast, standard LLMs achieve near-perfect accuracy (over 90\%) when given the textual content directly. This stark divergence between Text-QA(>90\%) and DeepSeek-OCR VQA(\textasciitilde20\%) proves that the information necessary for answering the questions exists in the document but DeepSeek-OCR's optical compression destroys the structured meaning required for reasoning.

Interestingly, even when DeepSeek-OCR's decoder is provided with uncompressed ground-truth text, it only achieves 27.7\% accuracy.
This suggests that the model may be over-optimized for surface-level text reconstruction at the expense of general linguistic reasoning capabilities. 

\subsection{Case Study}
Figure \ref{fig:case4} exemplifies this failure.
In a query regarding legal consequences ("contempt of court"), standard LLMs correctly deduce answer "B" from the text. DeepSeek-OCR, however, selects distinct incorrect answers in QA ("A") and VQA ("C") modes. 
This inconsistency highlights that the model is not grounding its answers in the document's content but is instead drifting primarily based on probability distributions or superficial associations, unanchored by the actual visual or textual evidence.


\section{RQ5: What Is the Context Length Limit for Optical Compression?}
The core premise of DeepSeek-OCR is that optical compression can bypass the quadratic complexity of standard LLMs, theoretically enabling infinite context windows via efficient visual tokens.
In this final analysis, we stress-test this claim. We investigate whether the "optical context" is truly scalable, or if the fixed resolution of vision encoders imposes a hard information-theoretic ceiling that triggers catastrophic model collapse.


\subsection{Experimental Setup}
To evaluate the limits of optical compression, we construct a controlled long-form narrative benchmark.
We prompt GPT-5.1 to generate five English stories with 5k words each.
To achieve long contexts, Each story is repeated until reaching approximately 20,000 tokens.
Each story is then segmented into 40 spans (500–20,000 tokens, 500-token steps) and rendered as document images.
Evaluation is performed in Tiny, Small, Base, and Large modes of DeepSeek-OCR to determine if scaling the visual encoder mitigates context length limitations.

\subsection{Results and Analysis}
\begin{figure}[htbp]
    \centering
    \includegraphics[width=0.48\textwidth]{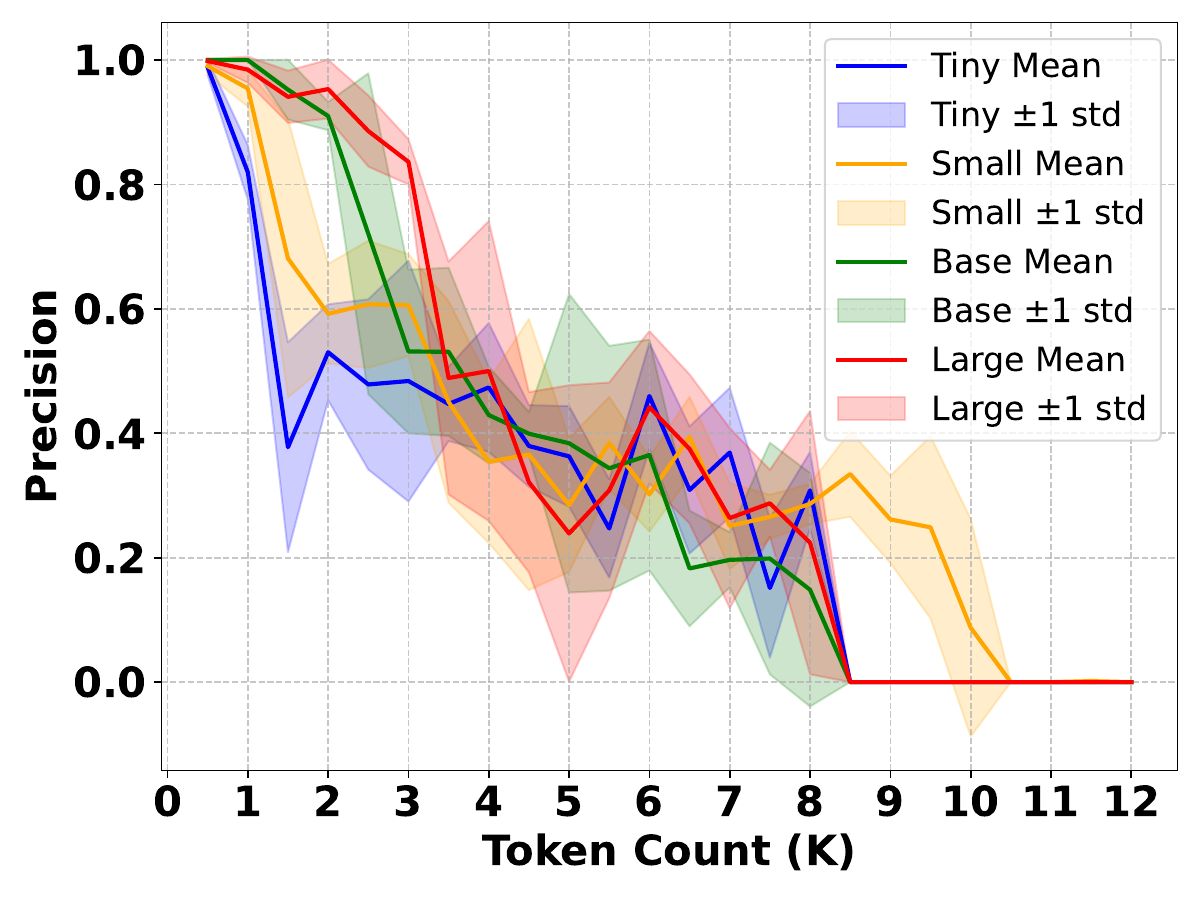}
    \caption{Compression and decompression results for different context lengths.}
    \vspace{-1em}
    \label{fig:compress}
\end{figure}

Figure ~\ref{fig:compress} plots the error band of OCR precision vs context length for each mode.
For the sake of presentation clarity and aesthetics, we truncated the image at the 12,000 token, since all subsequent data values were zero.
Contrary to the claims of handling long contexts, all DeepSeek-OCR modes exhibit a sharp performance cliff:
\begin{itemize}
    \item \textbf{The 8.5k Barrier:} Regardless of the compression modes, we find a systemic collapse point.
    The Tiny mode maintains viability only up to \textasciitilde6,000 tokens before plummeting to zero precision by 8,500 tokens.
    Surprisingly, scaling up to Base and Large modes yields diminishing returns, they also suffer complete collapse by 8,500 tokens.
    But there is a strange phenomenon that Small mode slightly outperforms Base and Large modes at extreme lengths (collapse at 10,500 tokens), possibly due to overfitting or instability in larger models under extreme compression.
    \item \textbf{The Density-Fidelity Trade-off:} These results suggest a fundamental limitation in the current paradigm: the amount of information a fixed-grid encoder can capture is finite.
    Once the text density exceeds this limit(\textasciitilde8.5k tokens per logical image unit), the signal-to-noise ratio drops below the decoder's recovery threshold, rendering the visual tokens meaningless.
\end{itemize}

Our stress test exposes a critical paradox: current optical compression techniques alleviate the computational bottleneck of processing tokens, but they introduce a far more restrictive information bottleneck.
With a hard ceiling around 10,000 tokens, DeepSeek-OCR effectively fails to handle the long contexts as it was designed to solve.




\subsection{Case Study}
Tab \ref{tab:case5} shows the progressive degradation of OCR results as context length increases from 2,500 to 8,500 tokens in DeepSeek-OCR Base mode.
At 2,500 tokens, the result almost perfectly reconstructs the original text, only with a minor error ("Xingfu" misrecognized as "Yingxiu").
However, with only increasing 500 tokens to 3,000, significant errors emerge, we can find the first half sentence is completely hallucinated and unrelated to the original text.
At 3,500 and 4,000 tokens, the OCR result is totally hallucinated by the decoder, with repeated phrases "The first time in the past" in 3,500 tokens and "narrow hallway with a long" in 4,000 tokens.
And by 8,000 tokens, the OCR result is simply an irrelevant sentence: "The following text is a placeholder for an image. It does not contain any relevant information for the article."
After 8,500 tokens, the model completely collapses and fails to produce any meaningful text, the OCR output completely consists of nonsense html tags: "<table><tr><td></td><td></td>..."
This case vividly illustrates the severe semantic degradation incurred through optical compression as context length increases.


\section{Related Works}
\subsection{Vision Encoders in Vision-Language Models}
VLMs employ three main vision encoder designs: dual-tower for parallel high-resolution processing but incur heavy preprocessing and training overhead \cite{vary}; tile-based for memory efficiency but produce excessive tokens \cite{internvl}; adaptive-resolution for flexibility but suffer quadratic memory growth \cite{qwen2.5-vl}. 
DeepSeek-OCR's DeepEncoder combines windowed SAM \cite{sam} and global CLIP \cite{clip} with a convolutional compressor, prioritizing low activations. 
But none of these designs isolate linguistic priors from visual recognition, leaving open how much OCR performance reflects true visual understanding.

\subsection{End-to-End OCR and Document Understanding Models}
End-to-end OCR has replaced traditional pipelines. Nougat \cite{nougat} pioneered paper parsing; GOT-OCR2.0 \cite{gotocr} broadened dense tasks; VLMs like Qwen-VL \cite{qwen2.5-vl}, InternVL \cite{internvl}, Donut \cite{donut}, and Pix2Struct \cite{pix2struct} enhanced OCR; specialized models like MinerU \cite{mineru}, UDOP \cite{udop}, and DocLLM \cite{docllm} target layouts.
Despite high accuracies, evaluations focus on edit-distance or ANLS on natural text, leaving unanswered how few vision tokens are needed for meaningful text decoding.

\subsection{Linguistic Priors in Multimodal Models}

Multimodal models frequently exploit language priors over visual signals. 
Blind LLMs can outperform vision-enabled ones on some VQA tasks\cite{lin2023revisiting}. 
CLIP shows textual biases\cite{luo2024probing,materzynska2022disentangling}.
In OCR, large models excel at printed text but struggle with handwritten or complex layouts, indicating reliance on linguistic context\cite{liu2024ocrbench}. 
Probing studies suggest high compressed OCR scores often reflect decoder hallucination\cite{laurenccon2024matters,luo2024probing}.

\section{Conclusion}

This paper provides an in-depth dissection of DeepSeek-OCR, revealing its performance relies heavily on linguistic priors rather than visual encoding. 
Our analysis shows that these priors artificially inflate accuracy by 60–80\% under compression; in zero-prior settings, performance precipitates to approximately 20\%. 
This dependency extends to end-to-end VLMs, a finding corroborated by sentence-level and word-level disruption tests. 
Unlike vision-centric models which demonstrate robustness, DeepSeek-OCR exhibits significant fragility: VQA tasks reveal a near-random semantic loss of \textasciitilde20\%, and long-context capabilities fail between 8,000–10,500 tokens. 
We conclude that current optical compression strategies prioritize token reduction at the expense of fidelity, rendering them inadequate for long-context applications without architectural redesign. 
Consequently, we advocate for prior-agnostic evaluation protocols—incorporating semantic disruptions and reasoning tasks—to guide the development of more robust systems.
\section*{Limitations}
Our current analysis primarily focuses on dense textual content to evaluate the vision-text compression paradigm. We have not extensively benchmarked the model's performance on structured or high-entropy data types, such as complex mathematical formulas, code snippets, or dense tabular data, where linguistic priors are naturally less predictive. It remains to be verified whether the observed "linguistic crutch" phenomenon persists to the same degree in these low-context scenarios.



\bibliography{ref}

@article{deepseek-ocr,
  title={Deepseek-ocr: Contexts optical compression},
  author={Wei, Haoran and Sun, Yaofeng and Li, Yukun},
  journal={arXiv preprint arXiv:2510.18234},
  year={2025}
}

@article{fox,
  title={Focus anywhere for fine-grained multi-page document understanding},
  author={Liu, Chenglong and Wei, Haoran and Chen, Jinyue and Kong, Lingyu and Ge, Zheng and Zhu, Zining and Zhao, Liang and Sun, Jianjian and Han, Chunrui and Zhang, Xiangyu},
  journal={arXiv preprint arXiv:2405.14295},
  year={2024}
}

@article{monkeyocr,
  title={MonkeyOCR: Document Parsing with a Structure-Recognition-Relation Triplet Paradigm},
  author={Li, Zhang and Liu, Yuliang and Liu, Qiang and Ma, Zhiyin and Zhang, Ziyang and Zhang, Shuo and Guo, Zidun and Zhang, Jiarui and Wang, Xinyu and Bai, Xiang},
  journal={arXiv preprint arXiv:2506.05218},
  year={2025}
}

@article{nougat,
  title={Nougat: Neural optical understanding for academic documents},
  author={Blecher, Lukas and Cucurull, Guillem and Scialom, Thomas and Stojnic, Robert},
  journal={arXiv preprint arXiv:2308.13418},
  year={2023}
}

@article{gotocr,
  title={General ocr theory: Towards ocr-2.0 via a unified end-to-end model},
  author={Wei, Haoran and Liu, Chenglong and Chen, Jinyue and Wang, Jia and Kong, Lingyu and Xu, Yanming and Ge, Zheng and Zhao, Liang and Sun, Jianjian and Peng, Yuang and others},
  journal={arXiv preprint arXiv:2409.01704},
  year={2024}
}

@article{paddleocr,
  title={Paddleocr 3.0 technical report},
  author={Cui, Cheng and Sun, Ting and Lin, Manhui and Gao, Tingquan and Zhang, Yubo and Liu, Jiaxuan and Wang, Xueqing and Zhang, Zelun and Zhou, Changda and Liu, Hongen and others},
  journal={arXiv preprint arXiv:2507.05595},
  year={2025}
}

@article{qwen2.5-vl,
  title={Qwen2. 5-vl technical report},
  author={Bai, Shuai and Chen, Keqin and Liu, Xuejing and Wang, Jialin and Ge, Wenbin and Song, Sibo and Dang, Kai and Wang, Peng and Wang, Shijie and Tang, Jun and others},
  journal={arXiv preprint arXiv:2502.13923},
  year={2025}
}

@article{minicpm,
  title={Minicpm: Unveiling the potential of small language models with scalable training strategies},
  author={Hu, Shengding and Tu, Yuge and Han, Xu and He, Chaoqun and Cui, Ganqu and Long, Xiang and Zheng, Zhi and Fang, Yewei and Huang, Yuxiang and Zhao, Weilin and others},
  journal={arXiv preprint arXiv:2404.06395},
  year={2024}
}

@misc{qwen3vl,
      title={Qwen3-VL Technical Report}, 
      author={Shuai Bai and Yuxuan Cai and Ruizhe Chen and Keqin Chen and Xionghui Chen and Zesen Cheng and Lianghao Deng and Wei Ding and Chang Gao and Chunjiang Ge and Wenbin Ge and Zhifang Guo and Qidong Huang and Jie Huang and Fei Huang and Binyuan Hui and Shutong Jiang and Zhaohai Li and Mingsheng Li and Mei Li and Kaixin Li and Zicheng Lin and Junyang Lin and Xuejing Liu and Jiawei Liu and Chenglong Liu and Yang Liu and Dayiheng Liu and Shixuan Liu and Dunjie Lu and Ruilin Luo and Chenxu Lv and Rui Men and Lingchen Meng and Xuancheng Ren and Xingzhang Ren and Sibo Song and Yuchong Sun and Jun Tang and Jianhong Tu and Jianqiang Wan and Peng Wang and Pengfei Wang and Qiuyue Wang and Yuxuan Wang and Tianbao Xie and Yiheng Xu and Haiyang Xu and Jin Xu and Zhibo Yang and Mingkun Yang and Jianxin Yang and An Yang and Bowen Yu and Fei Zhang and Hang Zhang and Xi Zhang and Bo Zheng and Humen Zhong and Jingren Zhou and Fan Zhou and Jing Zhou and Yuanzhi Zhu and Ke Zhu},
      year={2025},
      eprint={2511.21631},
      archivePrefix={arXiv},
      primaryClass={cs.CV},
      url={https://arxiv.org/abs/2511.21631}, 
}

@inproceedings{vary,
  title={Vary: Scaling up the vision vocabulary for large vision-language model},
  author={Wei, Haoran and Kong, Lingyu and Chen, Jinyue and Zhao, Liang and Ge, Zheng and Yang, Jinrong and Sun, Jianjian and Han, Chunrui and Zhang, Xiangyu},
  booktitle={European Conference on Computer Vision},
  pages={408--424},
  year={2024},
  organization={Springer}
}

@inproceedings{internvl,
  title={Internvl: Scaling up vision foundation models and aligning for generic visual-linguistic tasks},
  author={Chen, Zhe and Wu, Jiannan and Wang, Wenhai and Su, Weijie and Chen, Guo and Xing, Sen and Zhong, Muyan and Zhang, Qinglong and Zhu, Xizhou and Lu, Lewei and others},
  booktitle={Proceedings of the IEEE/CVF conference on computer vision and pattern recognition},
  pages={24185--24198},
  year={2024}
}

@inproceedings{sam,
  title={Segment anything},
  author={Kirillov, Alexander and Mintun, Eric and Ravi, Nikhila and Mao, Hanzi and Rolland, Chloe and Gustafson, Laura and Xiao, Tete and Whitehead, Spencer and Berg, Alexander C and Lo, Wan-Yen and others},
  booktitle={Proceedings of the IEEE/CVF international conference on computer vision},
  pages={4015--4026},
  year={2023}
}

@inproceedings{clip,
  title={Learning transferable visual models from natural language supervision},
  author={Radford, Alec and Kim, Jong Wook and Hallacy, Chris and Ramesh, Aditya and Goh, Gabriel and Agarwal, Sandhini and Sastry, Girish and Askell, Amanda and Mishkin, Pamela and Clark, Jack and others},
  booktitle={International conference on machine learning},
  pages={8748--8763},
  year={2021},
  organization={PmLR}
}

@inproceedings{donut,
  title={Ocr-free document understanding transformer},
  author={Kim, Geewook and Hong, Teakgyu and Yim, Moonbin and Nam, JeongYeon and Park, Jinyoung and Yim, Jinyeong and Hwang, Wonseok and Yun, Sangdoo and Han, Dongyoon and Park, Seunghyun},
  booktitle={European Conference on Computer Vision},
  pages={498--517},
  year={2022},
  organization={Springer}
}

@inproceedings{pix2struct,
  title={Pix2struct: Screenshot parsing as pretraining for visual language understanding},
  author={Lee, Kenton and Joshi, Mandar and Turc, Iulia Raluca and Hu, Hexiang and Liu, Fangyu and Eisenschlos, Julian Martin and Khandelwal, Urvashi and Shaw, Peter and Chang, Ming-Wei and Toutanova, Kristina},
  booktitle={International Conference on Machine Learning},
  pages={18893--18912},
  year={2023},
  organization={PMLR}
}

@article{mineru,
  title={Mineru: An open-source solution for precise document content extraction},
  author={Wang, Bin and Xu, Chao and Zhao, Xiaomeng and Ouyang, Linke and Wu, Fan and Zhao, Zhiyuan and Xu, Rui and Liu, Kaiwen and Qu, Yuan and Shang, Fukai and others},
  journal={arXiv preprint arXiv:2409.18839},
  year={2024}
}

@inproceedings{udop,
  title={Unifying vision, text, and layout for universal document processing},
  author={Tang, Zineng and Yang, Ziyi and Wang, Guoxin and Fang, Yuwei and Liu, Yang and Zhu, Chenguang and Zeng, Michael and Zhang, Cha and Bansal, Mohit},
  booktitle={Proceedings of the IEEE/CVF conference on computer vision and pattern recognition},
  pages={19254--19264},
  year={2023}
}

@inproceedings{docllm,
  title={Docllm: A layout-aware generative language model for multimodal document understanding},
  author={Wang, Dongsheng and Raman, Natraj and Sibue, Mathieu and Ma, Zhiqiang and Babkin, Petr and Kaur, Simerjot and Pei, Yulong and Nourbakhsh, Armineh and Liu, Xiaomo},
  booktitle={Proceedings of the 62nd annual meeting of the association for computational linguistics (volume 1: long papers)},
  pages={8529--8548},
  year={2024}
}

@article{lin2023revisiting,
  title={Revisiting the role of language priors in vision-language models},
  author={Lin, Zhiqiu and Chen, Xinyue and Pathak, Deepak and Zhang, Pengchuan and Ramanan, Deva},
  journal={arXiv preprint arXiv:2306.01879},
  year={2023}
}

@article{luo2024probing,
  title={Probing visual language priors in vlms},
  author={Luo, Tiange and Cao, Ang and Lee, Gunhee and Johnson, Justin and Lee, Honglak},
  journal={arXiv preprint arXiv:2501.00569},
  year={2024}
}

@inproceedings{materzynska2022disentangling,
  title={Disentangling visual and written concepts in clip},
  author={Materzy{\'n}ska, Joanna and Torralba, Antonio and Bau, David},
  booktitle={Proceedings of the IEEE/CVF Conference on Computer Vision and Pattern Recognition},
  pages={16410--16419},
  year={2022}
}

@article{liu2024ocrbench,
  title={Ocrbench: on the hidden mystery of ocr in large multimodal models},
  author={Liu, Yuliang and Li, Zhang and Huang, Mingxin and Yang, Biao and Yu, Wenwen and Li, Chunyuan and Yin, Xu-Cheng and Liu, Cheng-Lin and Jin, Lianwen and Bai, Xiang},
  journal={Science China Information Sciences},
  volume={67},
  number={12},
  pages={220102},
  year={2024},
  publisher={Springer}
}

@article{laurenccon2024matters,
  title={What matters when building vision-language models?},
  author={Lauren{\c{c}}on, Hugo and Tronchon, L{\'e}o and Cord, Matthieu and Sanh, Victor},
  journal={Advances in Neural Information Processing Systems},
  volume={37},
  pages={87874--87907},
  year={2024}
}

\appendix

\section{Appendix}
\label{sec:appendix}

\subsection{Cases Illustration for RQ3, RQ4 and RQ5}
\begin{figure*}[t]
    \centering
    \includegraphics[width=0.95\textwidth]{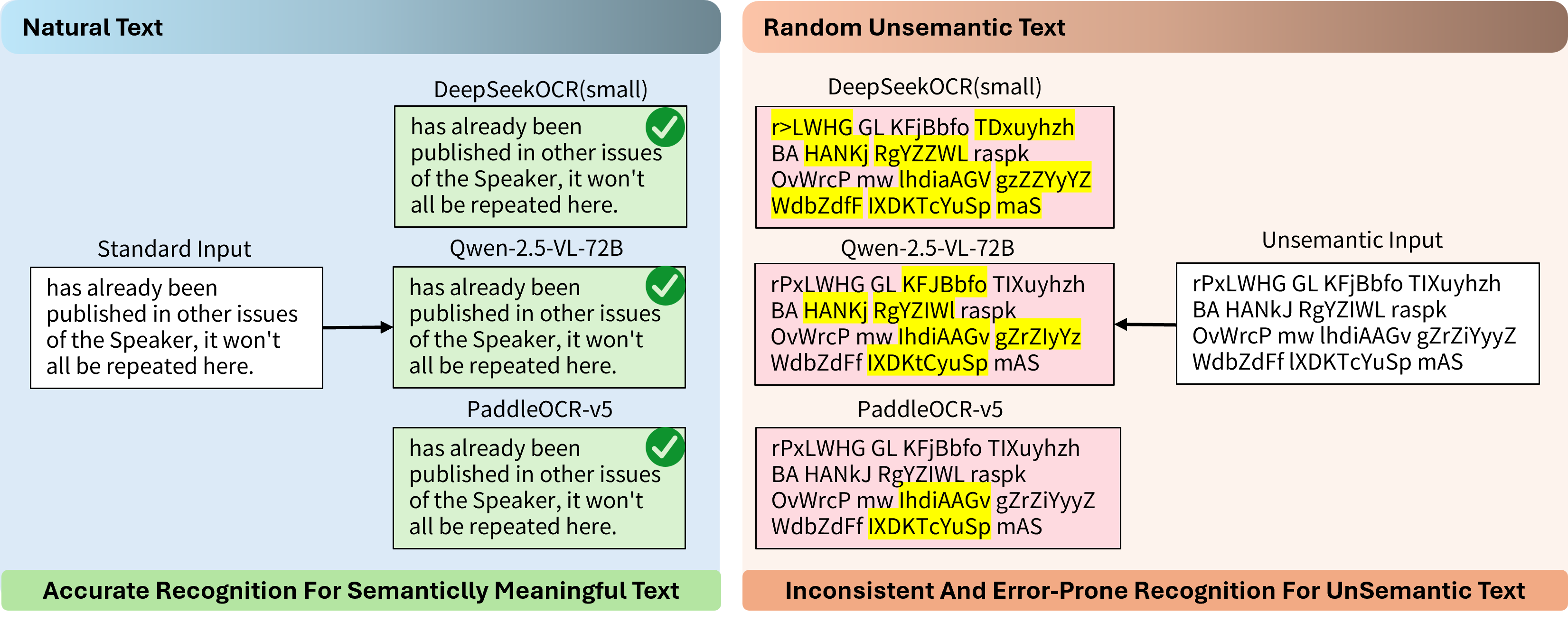}
    \caption{Case of OCR results on natural text and unsemantic text across different models.}
    \label{fig:case3}
\end{figure*}

\begin{figure*}[t]
    \centering
    \includegraphics[width=0.78\textwidth]{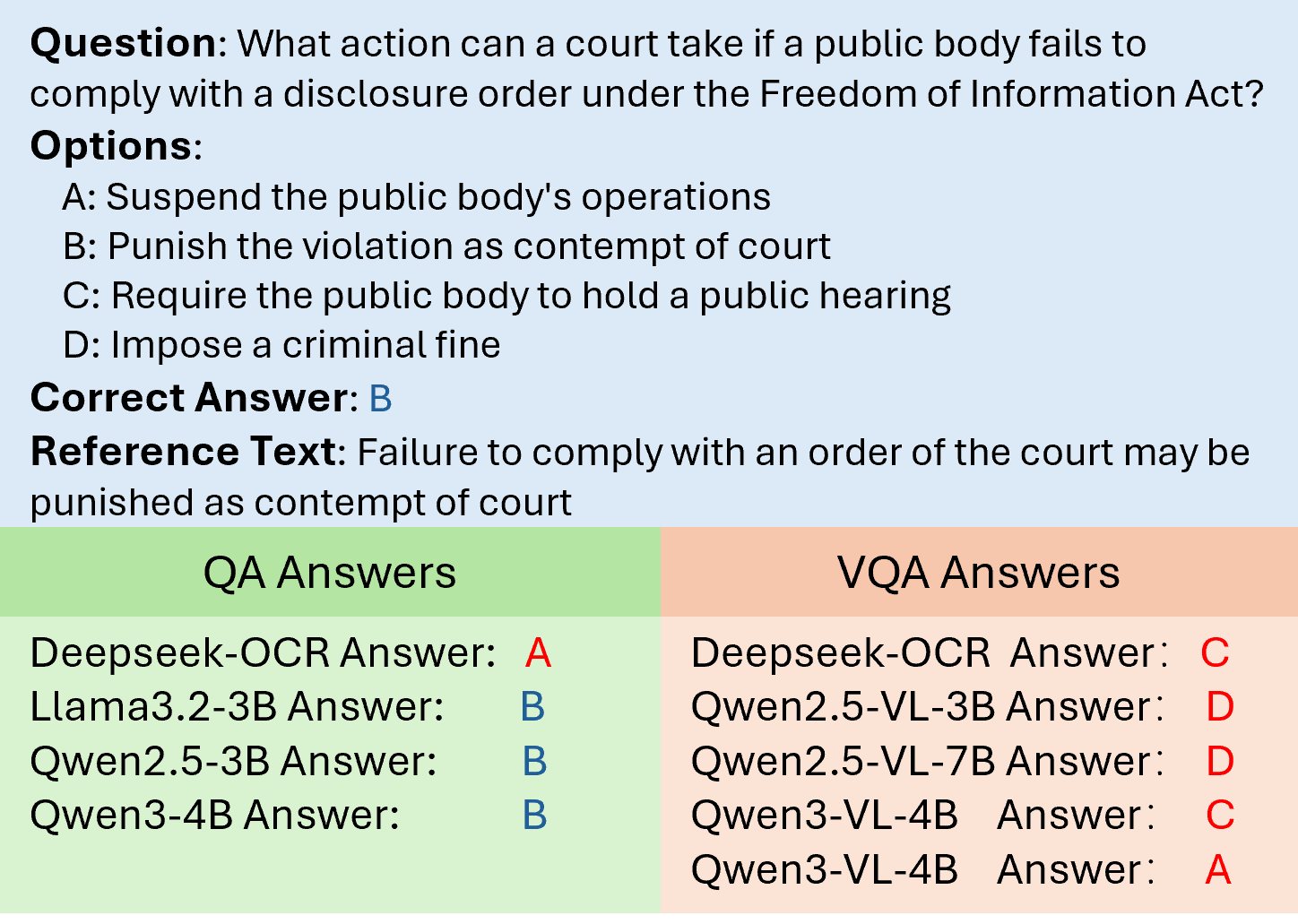}
    \caption{Case of QA and VQA results across different models.}
    \label{fig:case4}
\end{figure*}

\begin{table*}[t]
\centering
\small
\begin{tabularx}{\textwidth}{@{} l l X @{}}
\toprule
\textbf{Token Count} & \textbf{Category} & \textbf{Model Output Example \& Analysis} \\ \midrule

\rowcolor{gray!10}
Input Text & Ground Truth & The Old Phonograph. The first time Lin Mian saw the phonograph, it was tucked in the shadowy corner of the bookstore on Xingfu Road. \\ \midrule

\rowcolor{lightgreen}
2.5k & Almost Perfect & The Old Phonograph. The first time Lin \textcolor{red}{Man} saw the phonograph, it was tucked in the shadowy corner of the second-hand bookstore on \textcolor{red}{Yingxiu} Road. \\

\rowcolor{lightorange}
3k & Start Hallucination & \textcolor{red}{The growth of the Internet has run like a sawmilling machine.} It was \textcolor{red}{launched by the discoverer} of the second-hand bookstore on \textcolor{red}{Kingfisher} Road. \\

\rowcolor{lightorange}
3.5k & Repetitive & The world of the past. \textcolor{red}{The first time in the past. The first time in the past. The first time in the past...} (Repeated loops) \\

\rowcolor{lightorange}
4k & Repetitive & The ground floor of the train in San Mateo is the ground floor, it is located in the hallway of the car. The ground floor is a long, \textcolor{red}{narrow hallway with a long, narrow hallway with a long...} (Repeated loops) \\

\rowcolor{lightred}
8k & Irrelevant & \textit{``The following text is a placeholder for an image. It does not contain any relevant information for the article.''} \\

\rowcolor{red!20}
8.5k+ & \textbf{Model Collapse} & \texttt{<table><tr><td></td><td></td><td></td>...} (Structural breakdown into raw HTML/noise) \\ \bottomrule

\end{tabularx}
\caption{LLM Output Quality Degradation across Different Token Lengths}
\label{tab:case5}
\end{table*}

\end{document}